\documentclass{article}

\PassOptionsToPackage{numbers, compress}{natbib}
\usepackage[preprint]{neurips_2022}

\usepackage[utf8]{inputenc} 
\usepackage[T1]{fontenc}    
\usepackage[dvipsnames]{xcolor}
\usepackage[colorlinks=true,linkcolor=Purple,citecolor=Emerald]{hyperref}       
\usepackage{url}            
\usepackage{booktabs}       
\usepackage{amsfonts}       
\usepackage{nicefrac}       
\usepackage{microtype}      

\usepackage{amsthm}
\newtheorem{theorem}{Theorem}[section]

\newtheorem{proposition}{Proposition}[section]

\theoremstyle{definition}
\newtheorem{definition}{Definition}[section]

\usepackage{mathtools}
\usepackage{tikz} 

\newcommand{\av}[1]{}
\newcommand{\st}[1]{}
\newcommand{\ka}[1]{}
\newcommand{\guy}[1]{}
\newcommand{\kwc}[1]{}

\usepackage{graphicx}
\usepackage{notation}
\usepackage[capitalise,nameinlink]{cleveref}
\usepackage{xspace}

\newcommand{\BK}{\ensuremath{\mathsf{K}}\xspace}
\newcommand{\bTheta}{\ensuremath{\boldsymbol{\Theta}}\xspace}
\newcommand{\bbR}{\mathbb{R}}
\newcommand{\abs}[1]{|#1|}
\DeclareMathOperator*{\argmax}{argmax}

\newcommand{\semprola}{\textsc{SPL}\xspace}
\newcommand{\semprolas}{\textsc{SPL}s\xspace}
\newcommand{\semloss}{\ensuremath{\mathcal{L}_{\mathsf{SL}}}}

\newcommand{\sumUnit}{\textsc{Sum}}

\newcommand{\upparagraph}[1]{\vspace{-7pt}\paragraph{#1}}

\crefname{section}{Sec.}{Sec.}
\crefname{appendix}{Sec.}{Sec.}
\crefname{definition}{Def.}{Def.}

\usepackage{pifont}
\newcommand{\cmark}{\ding{51}}%
\newcommand{\xmark}{\ding{55}}%
\newcommand{\CMARK}{\textcolor{ForestGreen}{\cmark}}
\newcommand{\XMARK}{\textcolor{BrickRed}{\xmark}}

\usepackage{algorithm}
\usepackage[noend]{algorithmic}


\newcommand{\overparam}{\textsc{Overparameterize}}

\definecolor{ourspecialtextcolor}{rgb}{0.528, 0.471, 0.701}

\title{
Semantic Probabilistic Layers\\for Neuro-Symbolic Learning
}

\author{%
    Kareem Ahmed\\
    CS Department\\
    UCLA\\
    \texttt{ahmedk@cs.ucla.edu}\\
    \And
    Stefano Teso\\
    DISI\\
    University of Trento\\
    \texttt{stefano.teso@unitn.it}\\
    \And
    Kai-Wei Chang\\
    CS Department\\
    UCLA\\
    \texttt{kwchang@cs.ucla.edu}
    \AND
    Guy Van den Broeck\\
    CS Department\\
    UCLA\\
    \texttt{guyvdb@cs.ucla.edu}
    \And
    Antonio Vergari\\
    School of Informatics\\
    University of Edinburgh\\
    \texttt{avergari@ed.ac.uk}
}

\begin{document}
\maketitle
\begin{abstract}
    We design a predictive layer for structured-output prediction (SOP) that can be plugged into any neural network guaranteeing its predictions are consistent with a set of predefined symbolic constraints.
    Our {\bf S}emantic {\bf P}robabilistic {\bf L}ayer (\semprola) can model intricate correlations, and hard constraints, over a structured output space all while being amenable to end-to-end learning via maximum likelihood.
    \semprolas combine exact probabilistic inference with logical reasoning in a clean and modular way, learning complex distributions and restricting their support to solutions of the constraint.
    As such, they can faithfully, and efficiently, model complex SOP tasks beyond the reach of alternative  neuro-symbolic approaches.
    We empirically demonstrate that \semprolas outperform these competitors in terms of accuracy on challenging SOP tasks including hierarchical multi-label classification, pathfinding and preference learning, while retaining \textit{perfect} constraint satisfaction.
\end{abstract}

\section{Introduction}\label{sec:intro}
Modularity is among the major factors that propelled the Cambrian explosion of deep learning ~\citep{goodfellow2016deep}. 
By stacking multiple \textit{differentiable} layers together, practitioners are able to train deep classifiers in an end-to-end fashion with little effort.
However, despite its flexibility, \textit{this modular approach to learning does not guarantee that the predictions of these models conform to our expectations as to what makes sense}.
On the contrary, unconstrained deep classifiers are notorious for leading to predictions that are inconsistent with the logical constraints governing an underlying domain.

\begin{figure}[!t]
    \centering
    \setlength{\tabcolsep}{10pt}
    \begin{tabular}{cccc}
        \includegraphics[width=0.18\linewidth]{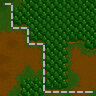}
        & \includegraphics[width=0.18\linewidth]{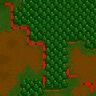}
        & \includegraphics[width=0.18\linewidth]{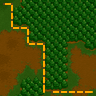}
        & \includegraphics[width=0.18\linewidth]{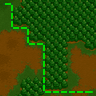}
        \\
        \textsc{Ground Truth} & \textsc{ResNet-18} & \textsc{Semantic Loss} & 
        \semprola (ours)
    \end{tabular}
    \caption{
    \textbf{Neural nets struggle with satisfying validity constraints in complex semantic SOP tasks}  such as predicting the lowest-cost path from the top-left to the bottom-right corners of  a Warcraft map.
        Even state-of-the-art neuro-symbolic approaches like the Semantic Loss~\cite{xu2018semantic} fail to ensure consistency with hard rules (c).
        \semprolas in contrast guarantees validity while retaining modularity, expressiveness and efficiency. 
        See \cref{sec:experiments} for complete experimental details and additional results. 
    } 
    \label{fig:illustration}
\end{figure}

This is even more evident in, and crucial for, structured output prediction (SOP) tasks, where classifiers have to predict hundreds of mutually constrained labels~\citep{tsochantaridis2004support,borchani2015survey}.
Consider for example a classical SOP task such as multi-label classification (MLC)~\citep{tsoumakas2007multi}.
Learning a multi-label classifier that disregards the correlations among labels, e.g., by considering them \textit{fully independent} given the inputs, yields sub-optimal results~\citep{bielza2011multi}.
In more challenging tasks such as hierarchical MLC (HMLC)~\citep{sorower2010literature} or pathfinding~\citep{poganvcic2019differentiation}, leveraging the domain's logical constraints (encoding, e.g., the label hierarchy or acyclicity and connectedness of a path) at training time can improve prediction accuracy~\citep{levatic2015importance}, but it cannot guarantee that the predictions are always \textit{consistent} with the constraints at inference time~\citep{giunchiglia2020coherent}.
\cref{fig:illustration} illustrates this problem in the context of pathfinding: constraint-unaware neural networks systematically fail to predict label configurations that form a valid path.
In many safety-critical scenarios such as protein function~\citep{radivojac2013large} and interaction prediction~\citep{sacca2014improved}, and drug discovery~\citep{de2018molgan,di2020efficient}, predicting inconsistent solutions can not only be harmful but also highly expensive~\citep{amodei2016concrete,giunchiglia2022deep}.

Unsurprisingly, due to their discrete nature, injecting logical constraints into deep neural networks while retaining modularity and differentiability is extremely challenging, as demonstrated in the \textit{neuro-symbolic learning} literature~\citep{sarker2021neuro}.
One such attempt has been to learn neural networks that satisfy the logical constraints by explicitly minimizing a differentiable loss term, the probability that the networks violates the constraint for any given prediction. And while successful, such approaches do not guarantee consistency of the predictions at test time.
More recently, researchers have proposed predictive layers that do guarantee consistency, but these are restricted to specific kinds of symbolic knowledge~\citep{giunchiglia2020coherent,sivaraman2020counterexample} or become intractable for even moderately complex logical constraints~\citep{Hoernle2021MultiplexNetTF}.

Motivated by these observations, we introduce a novel {\bf S}emantic {\bf P}robabilistic {\bf L}ayer (\semprola) for
modeling intricate correlations, and logical constraints on the labels of the output space in a modular and probabilistically sound manner.
It does so by leveraging recent advancements in the literature on probabilistic circuits~\citep{vergari2020probabilistic,choi2020pc}.
The key features of \semprola are that, on the one hand, it can be used as a \textit{drop-in replacement} for common predictive layers of deep nets like sigmoid layers, and on the other, it \textit{guarantees} the output's consistency with any prespecified logical constraints.
Importantly, \semprola also supports efficient inference and -- perhaps surprisingly -- does not complicate training.

\upparagraph{Contributions.}  Summarizing, we:
(\textit{i}) Identify six desiderata that neuro-symbolic predictors for constrained SOP tasks ought to satisfy, and show that state-of-the-art approaches fall short of one or more of them (\cref{tab:methods});
(\textit{ii}) Introduce \semprola, a novel semantic probabilistic layer that leverages probabilistic circuits to satisfy all six desiderata, i.e., that can be plugged into neural networks to ensure predictions to comply to given logical constraints, while retaining efficiency, expressivity, differentiability, and fully probabilistic semantics;
(\textit{iii}) We provide empirical evidence of the effectiveness of \semprola in several challenging neuro-symbolic SOP tasks, such as HMLC and pathfinding, where it outperforms state-of-the-art neuro-symbolic approaches, often by a noticeable margin.

\section{Designing a probabilistic layer for neuro-symbolic SOP}
\label{sec:design}

\upparagraph{Notation.}  In the following, we denote scalar constants $x$ in lower case, random variables $X$ in upper case, vectors of constants $\x$ in bold and vectors of random variables $\X$ in capital boldface.
$\Ind{\varphi}$ denotes the indicator function that evaluates to $1$ if the statement $\varphi$ holds and to $0$ otherwise.
We denote by $\x\models\BK$ that the value assignment $\x$ to variables $\X$ satisfies a logical formula \BK.

\upparagraph{Neuro-symbolic SOP.}  We tackle SOP tasks in which a neural net classifier must learn to associate instances $\x \in \bbR^D$ to $L$ \emph{interdependent} labels, identified by the vector $\y \in \{0, 1\}^L$.
We assume that we can abstract any neural classifier into two components:
 a feature extractor $f$ that maps inputs $\X$ to a $M$-dimensional embedding $\Z=f(\X)$ and a {predictive final layer}  that outputs the label distribution $p(\Y\mid\Z)$.
For example, the simplest, and yet widely adopted~\citep{mullenbach2018explainable,xu2018semantic,giunchiglia2020coherent}, predictive layer in neural classifiers for SOP considers labels $Y_i$ to be conditionally independent from each other given $\Z$, i.e., $p(\Y\mid\Z)=\prod_{i=1}^{L}p(Y_i\mid\Z)$.
We refer to this as \textit{fully independent layer} (FIL).
In a FIL, $p(Y_i = y_i\mid\z)$ is 
computed as 
$\sigma(\mathbf{w}_i^{\top} \z)$ where $\mathbf{w}_i\in\bbR^M$ is a vector of parameters and $\sigma(x)$ is the sigmoid function $1/(1+e^{x})$.

We are interested in dependencies between labels that can
occur both as \textit{correlations}, as is the case in MLC~\citep{dembczynski2012label}, and as \textit{logical constraints} 
encoded by logical formulas.
For example, in a HMLC task~\citep{giunchiglia2020coherent}
one logical constraint can encode the fact that observing a label for the class cat and dog, implies observing the label for their superclass animal
\begin{equation}
    \label{eq:label-logic}
    (Y_{\mathsf{cat}} \implies Y_{\mathsf{animal}}) \wedge (Y_{\mathsf{dog}} \implies Y_{\mathsf{animal}}).
\end{equation}

Specifically, we assume symbolic knowledge to be supplied in the form of  constraints encoded as a logical formula \BK
over the labels $\Y$ and optionally over a subset of the discrete input variables in $\X$, if any 
(e.g., in our experiments, the predicted simple path is constrained to lie within the subset of edges appearing in the input graph, see \cref{sec:grids}).
On the other hand, we expect a model to learn the label dependencies from data.
We call such tasks \textit{neuro-symbolic SOP }tasks.

\begin{table}[!t]
    \centering
    \setlength{\tabcolsep}{2pt}
    \caption{\textbf{\semprola is the only approach to satisfy all the desiderata for neuro-symbolic SOP.}
    An in-depth discussion of all competitors can be found in \cref{sec:related}.}
    \label{tab:methods}
    \scalebox{.82}{
    \begin{tabular}{lcccccccc}
        \toprule
            & \multicolumn{3}{c}{\sc Losses}
            & \multicolumn{5}{c}{\sc Layers}
        \\
        \cmidrule(lr){2-4} \cmidrule(lr){5-9}
        {\sc Desideratum}
            & {\sc DL2}~\cite{fischer2019dl2}
            & {\sc SL}~\cite{xu2018semantic}
            & {\sc NeSyEnt}~\cite{ahmed2022neuro}
            & {\sc FIL}
            & {\sc EBM}~\cite{lecun2006tutorial}
            & {\sc MultiplexNet}~\cite{Hoernle2021MultiplexNetTF}
            & {\sc CCN}~\cite{giunchiglia2021multi}
            & \semprola (Ours)
        \\
        \midrule
        (D1) Probabilistic  & \XMARK & \CMARK & \CMARK & \CMARK & \XMARK & \CMARK & \XMARK & \CMARK \\
        (D2) Expressive     & \XMARK & \XMARK & \XMARK & \XMARK & \CMARK & \XMARK & \XMARK & \CMARK \\
        (D3) Consistent     & \XMARK & \XMARK & \XMARK & \XMARK & \XMARK & \CMARK & \CMARK & \CMARK \\
        (D4) General        & \CMARK & \CMARK & \CMARK & \XMARK & \CMARK & \CMARK & \XMARK & \CMARK \\
        (D5) Modular        & \CMARK & \CMARK & \CMARK & \CMARK & \CMARK & \CMARK & \CMARK & \CMARK \\
        (D6) Efficient      & \CMARK & \CMARK & \CMARK & \CMARK & \XMARK & \XMARK & \CMARK & \CMARK \\
        \bottomrule
    \end{tabular}
    }
\end{table}

\upparagraph{Desiderata for neuro-symbolic SOP.}  To tackle this setting, we seek an algorithmic strategy for replacing the predictive layer in any neural network classifier with little effort, with the aim of injecting complex symbolic knowledge and allowing for flexible probabilistic reasoning.
We formalize these observations into the following six desiderata for our predictive layer:
\begin{itemize}

    \item[D1.] \textbf{Probabilistic}: The layer should enjoy sound probabilistic semantics, and deliver normalized probabilistic predictions to facilitate maximum-likelihood learning and sound decision making by virtue of calibrated probabilistic predictions
    
    \item[D2.] \textbf{Expressive}: It should be able to compactly encode intricate \textit{correlations} between labels.
    
    \item[D3.] \textbf{Consistent}:  It should always output predictions that are consistent with the prespecified symbolic knowledge, i.e., for all $\x$ and $\y$, if $(\x, \y) \not\models \BK$ then $p(\y\mid\x) = 0$.
    
    \item[D4.] \textbf{General}:  It should support rich \textit{logical constraints} over the labels expressed in some formal language, e.g., propositional logic.

    \item[D5.] \textbf{Modular}:  It should be applicable to any  off-the-shelf (and possibly pretrained) neural network in a modular fashion, enabling end-to-end learning and rapid prototyping.

    \item[D6.] \textbf{Efficient}:  The time required by the predictor to compute a prediction should be linear in the size of the predictor and of the hard constraint representation.

\end{itemize}

For example, FILs are clearly probabilistic (D1), modular (D5), and efficient (D6),
but at the cost of being incapable of modeling intricate correlations and logical constraints and thus generating inconsistent predictions (D2--D4) (see also \cref{fig:illustration}).
\cref{tab:methods} summarizes how the other popular and effective approaches to neuro-symbolic SOP nowadays fall short of one of more desiderata as well.
We discuss this in detail in \cref{sec:related}.
To the best of our knowledge, our proposed \textit{semantic probabilistic layers} (\semprolas) are the first algorithmic solution to satisfy all above desiderata.

\upparagraph{SPL.}
At a high level, \semprola realizes the above desiderata in a single layer 
that combines exact probabilistic inference with logical reasoning in a clean and modular way, learning complex distributions and restricting their support to solutions of the constraint.

\begin{figure}[!t]
    \centering
    \includegraphics[width=.27\columnwidth,page=2,trim=15 200 360 5, clip]{figs/semprola-graphics}\hfill
    \includegraphics[width=.27\columnwidth,page=3,trim=15 200 380 5, clip]{figs/semprola-graphics}\hfill
    \includegraphics[width=.4\columnwidth,page=1,trim=95 160 120 30, clip]{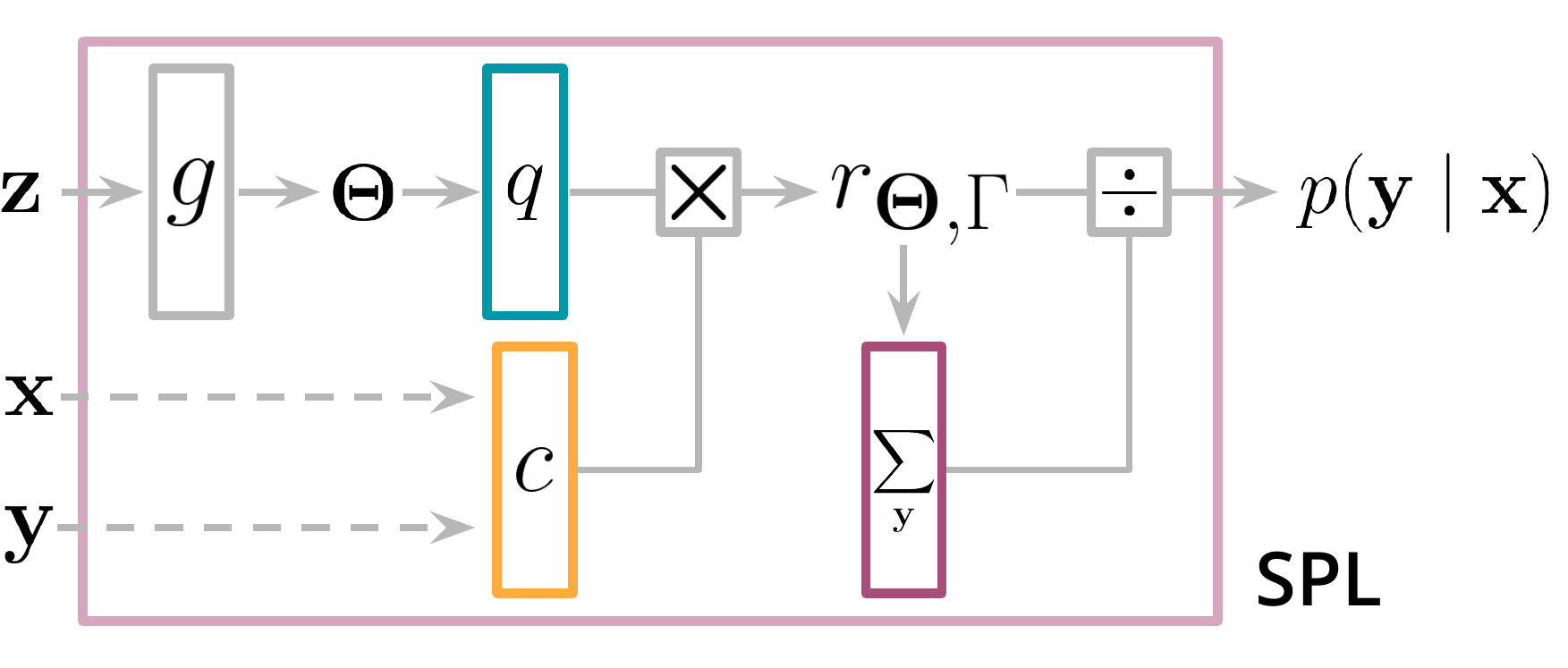}\\[-5pt]
    \caption{
    \textbf{A high level view of \semprolas.}
    The predictive layer of a neural network for neuro-symbolic SOP, e.g., a FIL (\textbf{left}), can be readily replaced by a \semprola (\textbf{middle}).
    \semprolas are implemented (\textbf{right}) by multiplying together a probabilistic circuit $q_{\bTheta}(\Y\mid f(\X))$ parameterized by (a function $g$ of) the network's embeddings $f(\X)$, and a constraint circuit $c_{\BK}(\X, \Y)$ embodying the symbolic knowledge.  The result is normalized by efficiently marginalizing over the product circuit $r_{\bTheta,\BK}$, so as to guarantee fully probabilistic semantics and end-to-end differentiable learning by maximum likelihood. 
    }
    \label{fig:semprola}
\end{figure}

\begin{definition}[Semantic probabilistic layer (SPL)]
  Given an input configuration $\x$, a SPL decomposes the computation of the probability of a label configuration as:
  \begin{equation}
                p(\y\mid f(\x))=q_{\bTheta}(\y\mid f(\x))\cdot c_{\BK}(\x, \y)\slash
                  \mathcal{Z}(\x)
                  \qquad
                  \text{where}
                  \quad
                  \mathcal{Z}(\x) = \sum\nolimits_{\y}q_{\bTheta}(\y\mid \x)\cdot c_{\BK}(\x, \y).
    \label{eq:semprola}
\end{equation}
Here, $q_{\bTheta}(\y\mid f(\x))$ is a module to perform probabilistic reasoning by encoding an expressive distribution over the labels parameterized by $\bTheta$; 
$c_{\BK}(\x, \y)$ is a module to ensure consistency of the predictions by encoding logical constraints $\BK$ and being non-zero only when $\BK$ is satisfied, i.e., $c_{\BK}(\x, \y)=\Ind{(\x, \y)\models\BK}$;
and $\mathcal{Z}(\x)$ is a renormalization term, also called the partition~function.
\end{definition}
\cref{fig:semprola} illustrates the computational graph of our \semprola at training time.
In order to satisfy all D1-D6, we will realize both $q_{\bTheta}$ and $c_{\BK}$ 
as \textit{circuits}~\citep{vergari2020probabilistic,choi2020pc}, constrained computational graphs that enable tractable computations.
Differently from FILs, $q_{\bTheta}$ in SPLs can encode an expressive joint distributions over the labels and therefore attain full expressiveness by scaling the number of parameters $\bTheta$ (D2).
Consistency is guaranteed by the component $c_{\BK}$:
by multiplying it to the joint probability of a label configuration  the resulting product distribution $r_{\bTheta,\BK}(\y, \x)=q_{\bTheta}(\y\mid f(\x))\cdot c_{\BK}(\x, \y)$ will have its support effectively cut by $\BK$, and thus cannot allocate any probability mass to inconsistent predictions (D3).
Additionally, $c_{\BK}$ will allow to encode general propositional logical constraints in a compact computational graph (D4).
Lastly, the product $r_{\bTheta,\BK}(\x, \y)$ is fully differentiable
and allows \semprola to be an off-the-shelf replacement for other predictive layers (see \cref{fig:semprola}) and enables end-to-end learning (D5).
By renormalizing $r_{\bTheta,\BK}(\x, \y)$
and outputting normalized probabilities, \semprola enables the exact computation of gradients for $\bTheta$, which can therefore be trained by maximum likelihood.

Thanks to recent advancements in the literature on circuits, 
we can compute the partition function $\mathcal{Z}(\x)$  efficiently in time linear in the size of  $r_{\bTheta,\BK}$, thus preserving efficiency (D6) and not compromising on the other desiderata.
This will also yield correct (and consistent) predictions at test time, when an SPL computes the MAP state $\y^{*}=\argmax_{\y}r_{\bTheta,\BK}(\y, \x)\slash \sum_{\y}r_{\bTheta,\BK}(\y, \x)$.
The next section clarifies \textit{how} to implement the modules of \semprola as circuits while satisfying these desiderata.

\section{Realizing \semprolas with tractable circuit representations}\label{sec:semprola}

The components of \semprolas are \textit{circuits}, a large class of computational graphs that can represent both functions and distributions~\citep{choi2020pc,darwiche2002knowledge}.
Circuits subsume many tractable generative and discriminative probabilistic models---from Chow-Liu and latent tree models~\citep{chow1968approximating,choi2011learning}, to hidden Markov models (HMMs)~\citep{rabiner1986introduction}, sum-product networks (SPNs)~\citep{poon2011sum},
decision trees~\citep{khosravi2020handling,correia2020joints}, and deep regressors~\citep{khosravi2019tractable}---as well as many compact representations of logical formulas, such as (ordered) binary decision diagrams~\citep{akers1978binary}, sentential decision diagrams (SDDs)~\citep{darwiche2011sdd} and others~\citep{darwiche2002knowledge}.

The key idea behind \semprolas is to leverage this single formalism to represent both an expressive joint distribution for $q_{\bTheta}(\y\mid f(\x))$ and a compact encoding of the logical constraints for $c_{\BK}(\x, \y)$, while ensuring the exact and efficient evaluation of \cref{eq:semprola}.
This can be achieved by ensuring that these computational graphs abide certain structural properties:
\textit{smoothness}, \textit{decomposability}, \textit{determinism} and \textit{compatibility}~\citep{darwiche2002knowledge,vergari2021compositional}.
Next, we introduce \textit{probabilistic circuits} for modeling $q_{\bTheta}$ (\cref{sec:prob-circuits})  and \textit{constraint circuits} for $c_{\BK}$ (\cref{sec:constraint-circuits}),
while in \cref{sec:single-semprola} we propose a more efficient implementation of \semprola utilizing a single circuit.

\subsection{Representing expressive distributions with probabilistic circuits}
\label{sec:prob-circuits}

We start by introducing circuits for \textit{joint} probability distributions, and then extend the discussion to \textit{conditional} distributions, which we use to implement $q_{\bTheta}(\Y\mid f(\X))$ in \semprolas.

\begin{definition}[Circuits]
\label{def:circuits}
A circuit $h$ over variables $\Y$ is a computational graph encoding a parameterized function $h_{\bTheta}(\Y)$  by combining three kinds of computational units: input functional units, sum units, and product units.
An input functional $n$ represents a base parametric function $h_n(\scope(n); \boldsymbol{\lambda})$ over some variables $\scope(n) \subseteq \Y$, called its scope, and it is parameterized by $\boldsymbol{\lambda}$.
Sum and product units $n$ elaborate the output of other units, denoted $\ch(n)$.
Sum units are parameterized by $\boldsymbol{\omega}$ and compute the weighted sum of their inputs $\sum_{c\in\ch(n)} \omega_{c} h_c(\scope(n))$,
while product units compute $\prod_{c\in\ch(n)} h_c(\scope(n))$.
The parameters $\bTheta$ of a circuit encompass the parameters of all input functionals ($\boldsymbol{\lambda}$) and sum units ($\boldsymbol{\omega}$).
\end{definition}

\begin{figure}[!t]
    \hspace{40pt}\includegraphics[height=.16\columnwidth,page=8]{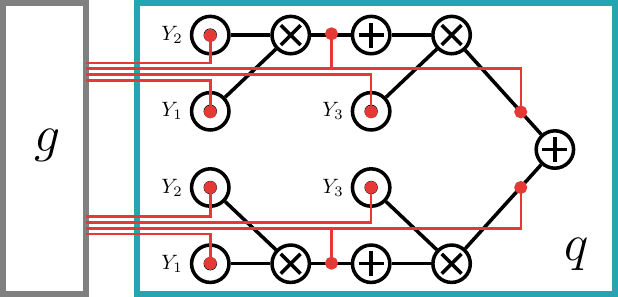}\hspace{30pt}
    \includegraphics[height=.16\columnwidth,page=7]{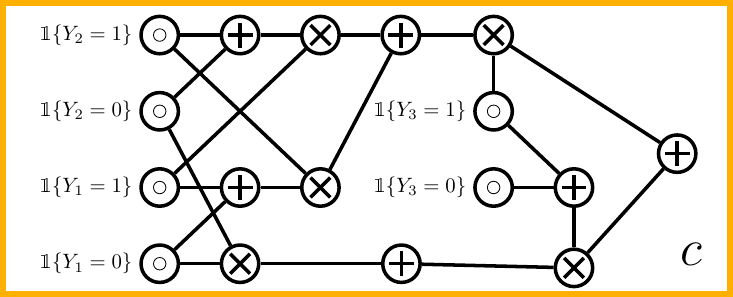}\hspace{20pt}
    \caption{
    \textbf{Examples of circuits in \semprola}.
    \textbf{Left}: probabilistic circuit $q_{\boldsymbol{\Theta}}$.  Red lines indicate how the parameters $\boldsymbol{\Theta}$ flow from the gating function $g$ to the various input functional and sum units of $q$.
    \textbf{Right}: constraint circuit encoding the logical constraint of \cref{eq:label-logic} where labels are $Y_i\in\{Y_{\mathsf{cat}},Y_{\mathsf{dog}}, Y_{\mathsf{animal}}\}$.
    Note that $q$ and $c$ are smooth, decomposable (\cref{def:smo-dec}) and compatible (\cref{def:compatibility}) and $c$ is deterministic (\cref{def:determinism}). By parameterizing $c$ via $g$ we can obtain a single-circuit \semprola (\cref{sec:single-semprola}).}
    \label{fig:semprola-circuits}
\end{figure}

For any input $\y$, the value of $h_{\bTheta}(\y)$ can be evaluated by propagating the output of the input units through the computational graph and reading out the value of the last unit.
The \emph{support} of $h$ is the set of all
states $\y$
for which the output
is non-zero, i.e., $\supp(h)=\{\y\in\val(\Y)\,|\, h(\y)\neq0\}$.

\begin{definition}[Probabilistic circuits (PCs)]
\label{def:prob-circuits}
A circuit $q$ is a PC if it encodes a (possibly unnormalized) probability 
distribution, i.e., $q_{\bTheta}(\y)$ is non-negative for all configurations $\y$ of $\Y$.
\end{definition}

From here on, we will assume PCs to have positive sum parameters $\boldsymbol{\omega}$ and whose input units model valid distributions, e.g., Bernoullis, as these conditions are sufficient for satisfying \cref{def:prob-circuits}.
Moreover, w.l.o.g.\ we will assume the sum and product units to be organized into alternating layers, and that every product unit $n$ receives only two inputs $c_1, c_2$, \ie $q_n(\X)=q_{c_1}(\Y)\cdot q_{c_2}(\Y)$.
These conditions can easily be enforced in polynomial time~\citep{vergari2015simplifying,vergari2019visualizing}.
We are specifically interested in smooth and decomposable PCs, as they 
will be enabling efficient inference in \semprola (\cref{sec:efficient-inference-semprolas}).

\begin{definition}[Smoothness \& Decomposability]
\label{def:smo-dec}
A circuit is \textit{smooth} if for every sum unit $n$, its inputs depend on the same variables: $\forall\, c_1, c_2\in\ch(n), \scope(c_1)=\scope(c_2)$.
It is \textit{decomposable} 
if the inputs of every product unit $n$ depend on disjoint sets of variables: 
$\ch(n)=\{c_1,c_2\}, \scope(c_1)\cap\scope(c_2)=\emptyset$.
\end{definition}

Smooth and decomposable PCs are both expressive and efficient:  they can encode distributions with hundred millions of parameters and be effectively learned by gradient ascent~\citep{peharz2020random}.
The structure of their computational graph can be either specified
manually~\citep{poon2011sum,peharz2020random,peharz2020einsum} or
acquired automatically from data~\citep{vergari2015simplifying,rahman2014cutset,dang2022strudel}, e.g., by first learning a latent tree model and then compiling the latter into a circuit~\citep{liu2021tractable}.
These circuits are competitive with intractable models such as variational autoencoders and normalizing flows scores on several benchmarks~\citep{liu2022lossless}.

As proposed by~\citet{shao2022conditional}, any (smooth and decomposable) PC $q_{\bTheta}(\Y)$ encoding a \textit{joint} distribution over the labels $\Y$ can be turned into a (smooth and decomposable) \textit{conditional} circuit, conditioned by input variables $\X$, by letting its parameters be a function of $\X$.

\begin{definition}[Neural conditional circuits~\citep{shao2020conditional}]
\label{def:cond-circuit}
A conditional circuit $q(\Y; \bTheta=g(\X))$ models the conditional distribution $p(\Y\mid \X)$ via a differentiable gating function $g$ that maps every input configuration $\x$ to the set of parameters of $\bTheta$ of $p$.
\end{definition}

An example of a smooth and decomposable conditional circuit is shown in \cref{fig:semprola-circuits}.
This design immediately allows us to implement $q_{\boldsymbol{\Theta}}(\Y\mid f(\X))$ in \semprola as a conditional PC whose gating function maps the feature embedding space $\mathbb{R}^{K}$ to the parameter space $\mathbb{R}_{+}^{|\bTheta|}$, realizing $q(\Y; \bTheta=g(f(\X)))$.
As such, the gating function $g$ creates a clean interface between any pre-trained feature extractor $f$ and the PC $q$ (\cref{fig:semprola-circuits}).
While one can devise $g$ in several ways, we strive for simplicity in our experiments and adopt vanilla multi-layer perceptrons (MLPs) whose final activations are either sigmoids, if they have to predict the parameters $\boldsymbol{\lambda}$ of the Bernoulli input distributions of $q$, or softmax, if they output the sum unit parameters $\boldsymbol{\omega}$ (\cref{def:circuits}).

\subsection{Encoding logical formulas with constraint circuits}
\label{sec:constraint-circuits}

The next step is to translate a logical constraint $\BK$ into a smooth and decomposable circuit $c_{\BK}(\x, \y)$.  
To this end, we employ a special type of PCs, defined as follows.

\begin{definition}[Constraint circuits]
\label{def:constraint-circuits}
A PC $c$ over variables $\X\cup\Y$ is a constraint circuit encoding prior knowledge $\BK$ if it computes $\Ind{(\x,\y)\models\BK}$ for every configuration $(\x,\y)$.
\end{definition}

As a practical way to realize such a circuit, we will consider constraint circuits that have all sum unit parameters equal to 1 and input functionals that are indicator functions over their scope, e.g., $c_n(\z)=\Ind{\z\models{\varphi(n)}}$ where $\Z$ is the scope of the input and $\varphi(n)$ a constraint over it. 
Furthermore, we require each sum unit in it to be \textit{deterministic}.
\begin{definition}[Determinism]
\label{def:determinism}
A sum unit $n$ is \textit{deterministic} if its inputs have disjoint supports, i.e., $\forall\, c_1,c_2\in\ch(n), c_1\neq c_2 \implies \supp(c_1)\cap\supp(c_2)=\emptyset$.
\end{definition}
\cref{fig:semprola-circuits} shows an example of a deterministic constraint circuit.
Thanks to determinism, we can readily translate 
classical compact representations for logical formulas such as (ordered) binary decision diagrams~\citep{akers1978binary,bryant2002ordered} and sentential decision diagrams (SDDs)~\citep{darwiche2011sdd} into constraint circuits as defined above.
This becomes evident when they are written in the language of negation normal form~\citep{darwiche2002knowledge} and their \textit{and} gates (resp. \textit{or} gates) are replaced with product units (resp. sum units)~\citep{choi2020pc}.
A logic constraint can therefore be represented as a constraint circuit for \semprolas, by utilizing any of the many tools available for OBDDs ~\citep{toda2016implementing} or SDDs~\citep{choi2013dynamic, oztok2015topdown}. 
\cref{app:compilation} illustrates in detail how to compile the example constraint of \cref{eq:label-logic} into the constraint circuit of \cref{fig:semprola-circuits} in this way.

\subsection{Efficient inference in \semprolas}
\label{sec:efficient-inference-semprolas}

As discussed above, PCs can be expressive (D1) and are modular (D5), while constraint circuits ensure consistency (D3) for general constraints (D4).
What remains to be shown to complete \semprolas is that the product supports efficient normalization (D2) and inference (D6), specifically that it allows for the efficient evaluation of the normalization constant of $r_{\bTheta,\BK}$, and its MAP state.
To this end, we need to introduce the notion of compatibility between the two circuits~\citep{vergari2021compositional}.

\begin{definition}[Compatible circuits in \semprolas]
    \label{def:compatibility}
    A smooth and decomposable conditional PC $q(\Y; \bTheta)$ is compatible over variables $\Y$  with a smooth and decomposable constraint circuit $c_{\BK}(\Y, \X)$  if any pair of product units $n\!\in\!\p$ and $m\!\in\! m$ with the same scope over $\Y$ can be rearranged to be mutually compatible and decompose in the same way:
    $(\scope(n)\!=\!\scope(m)) \!\implies\! (\scope(n_i)\!=\!\scope(m_i), \text{ $n_i$ and $m_i$ are compatible}) $ 
    for some rearrangement of the inputs of $n$ (resp.\ $m$) into $n_1,n_2$ (resp.\ $m_1,m_2$). \cref{fig:semprola-circuits} shows two circuits $q$ and $c$ that are compatible.
\end{definition}

\begin{theorem}[Efficient inference in \semprolas]
\label{thm:prod-pcs}
If $q(\Y; \bTheta)$ and $c_{\BK}(\Y, \X)$ are two smooth, decomposable and compatible circuits, then computing \cref{eq:semprola} can be done in $\mathcal{O}(\abs{q}\abs{c})$ time.
Furthermore, if they are also deterministic, then computing the MAP state can be done in $\mathcal{O}(\abs{q}\abs{c})$ time. 
\end{theorem}

The proof can be found in \cref{app:proofs}.
How do we come up with compatible circuits? 
One option is to have a PC $q$ that is compatible with every possible smooth and decomposable circuit $c$.
To do so, we can represent $q$ as a mixtures of $M$ fully-independent models; \ie 
$\sum_{i=1}^{M}\omega_i\prod_{j}q(Y_j; \bTheta_i)$.
This additional sum unit can be enough to be more expressive than a FIL and already delivers more accurate predictions than any competitor, as our experiments in pathfinding show (\cref{sec:experiments}). 
An example of such a circuit is shown in \cref{fig:semprola-circuits}.
Another sufficient condition for compatibility is that both $q$ and $c$ share the exact same hierarchical scope partitioning~\citep{vergari2021compositional}, sometimes called a vtree or variable ordering~\citep{choi2020pc,pipatsrisawat2008new}.
This can be done easily if one first compiles logical constraints into OBDDs or SDDs and then uses a mechanized algorithm to build $q$ as in \citep{peharz2020random} to create a compatible structure.
Additionally, to ensure $q$ is a deterministic PC, we could exploit the mechanized construction proposed in \citet{shih2020probabilistic}. 
Computing the exact MAP state, however, is of less concern as approximate inference algorithms, e.g., beam search decoding~\citep{vijayakumar2016diverse} or iterative pruning~\cite{choi2022solving}, are nowadays a commodity in deep learning frameworks.
For non-deterministic PCs, we compute the MAP state with a faster approximation by replacing non-deterministic sum units with max units~\cite{peharz2016latent}.
This runs in time linear in the size of $r$, and yet delivers state-of-the-art accuracies in our experiments~\cref{sec:experiments}.

\subsection{A single-circuit \semprola}
\label{sec:single-semprola}
The two-circuit design we proposed for \semprolas provides a clear and theoretically-backed interface between neural networks and probabilistic and symbolic reasoning.
This setup, however, can sometimes be wasteful, as it requires to compute the product of two circuits and renormalize.
We circumvent this issue by designing a single-circuit implementation of \semprola.
\begin{definition}[Single-circuit \semprola]
\label{def:single-circuit-semprola}
Given an input configuration $\x$, a single-circuit SPL computes $p(\y\mid\x)=c_{\BK}(\Y, \X; \boldsymbol{\Omega}=g(f(\X)))$ where $c_{\BK}$ is a neural conditional constraint circuit whose sum-unit parameters $\boldsymbol{\Omega}$ are non-unitary values parameterized via a gating function $g$.
\end{definition}
In a nutshell, we can directly realize \semprola by compiling a complex logical constraints (D4) into a deterministic constraint circuit $c_{\BK}$, as before, and then parameterizing 
it with a gating function
of the network embeddings $f(\X)$, i.e., allowing its sum units to be non-unitary and input dependent.
Since the support of $c_{\BK}$ is already restricted to exactly match the constraint $\BK$ (D3), parameterizing $\boldsymbol{\Omega}$ induces an expressive probability distribution over the label configurations that are consistent with $\BK$ (D2). 
We can further guarantee that the circuit's output are normalized probabilities (D1,D6) by enforcing the parameters $\omega$ of each sum unit to form a convex combination~\citep{peharz2015theoretical}.
This can be easily done by utilizing a softmax activation function for $g$.

One of the advantages of the two-circuit implementation of \semprolas is that the size of the circuit $q_{\boldsymbol\Theta}$ can be easily increased to improve the capacity of the model (\cref{sec:prob-circuits}).
The single-circuit implementation is not as flexible, as normally the number of parameters is determined by the complexity of the constraint circuit, which depends entirely on the compilation step.
In this case, one option is to \emph{overparameterize} the neural conditional circuit by introducing additional sum units, hence allowing it to capture more modes in the distribution.  
We detail this process is \cref{app:overparameterize}.
A side effect of overparameterization is that it relaxes determinism, meaning that MAP inference needs to be approximated, as described in \cref{sec:efficient-inference-semprolas}.
Additionally, training a gating function to map relatively small embeddings to large parameter vectors in overparameterized circuits, can slow down training.
In such cases, a two-circuit implementation of \semprola is to be preferred.

\begin{table}[!t]
\centering
\caption {\semprolas outperform all loss-based competitors in the neuro-symbolic benchmarks of \citep{xu2018semantic}.}
\vskip -0.05in
\scalebox{.8}{
{\sc
\begin{tabular}{@{}l r r r r r r}
\toprule
& \multicolumn{3}{c}{Simple Path} & \multicolumn{3}{c}{Preference Learning}\\
\cmidrule(lr){2-4} \cmidrule(lr){5-7}
Architecture  & Exact & Hamming & Consistent & Exact & Hamming & Consistent\\
\midrule
MLP+FIL &   5.6 & 85.9  & 7.0 &   1.0 & \textbf{75.8}  & 2.7\\
MLP+\semloss& 28.5 &  83.1 & 75.2 & 15.0 &  72.4 & 69.8 \\
MLP+NeSyEnt& 30.1 &  83.0 & 91.6 & 18.2 &  71.5 & 96.0 \\
MLP+\semprola& \textbf{37.6} & \textbf{88.5} & \textbf{100.0}
& \textbf{20.8} & 72.4 & \textbf{100.0}
\\
\bottomrule
\end{tabular}
}
}
\label{tab:sp}
\end{table}

\section{Related works}
\label{sec:related}

In this section, we position \semprolas against state-of-the-art approaches for enforcing constraints on neural network predictions.  In-depth surveys on this topic can be found in~\citep{dash2022review} and~\citep{giunchiglia2022deep}.

\noindent
\textbf{Energy-based models.}  Deep energy-based models (EBMs) replace FILs with an unnormalized factor graph~\citep{koller2009probabilistic} that captures higher-order label dependencies~\cite{lecun2006tutorial} (D2) but at the cost of foregoing probabilistic semantics (D1) and efficiency (D6). 
EBMs are typically unconcerned with hard constraints (D3).
Neural approaches for segmentation \citep{liu2015crf} and parsing \citep{DBLP:conf/acl/DurrettK15,DBLP:conf/acl/ZhangLZ20,DBLP:conf/ijcai/ZhangZL20} remedy to this by replacing the factor graph with a full-fledged intractable (discriminative) graphical model~\citep{koller2009probabilistic}.  
To gain efficiency, one can restrict EBMs to simpler graphical models (e.g., chains, trees), compromising expressiveness (D2) and the ability to model non-trivial logical constraints (D3, D4).

\noindent
\textbf{Loss-based methods.}  A prominent strategy consists of penalizing the network for producing inconsistent predictions using an auxiliary loss~\citep{dash2022review,giunchiglia2022deep}.
While popular, loss-based methods, however \textit{cannot} guarantee that the predictions will be consistent at test time.
Common losses include translating logical constraints into a differentiable fuzzy logic~\citep{diligenti2012bridging,diligenti2017semantic}, as exemplified by DL2~\cite{fischer2019dl2}.
Although efficient (D6), this solution is not probabilistically sound (D1) and crucially \textit{is  not syntax-invariant}: different encodings of the same formula (e.g., conjunctive vs.\@ disjunctive normal form) yield different losses~\cite{giannini2018convex,di2020efficient}.
Closer to our \semprola, the Semantic Loss (SL)~\citep{xu2018semantic} avoids this issue by
penalizing the the probability $\theta_i$ associated to the $i$-th label by the neural network via the loss term
\begin{equation*}
        \semloss \propto -\sum_{y\models\BK}\prod_{\y\models Y_i}\theta_i\prod_{\y\not\models Y_i}(1-\theta_i) =  -\sum_{\y\models\BK} \prod_i p(Y_i\mid\x) = -\sum_{\y}\prod_i  q(Y_i; \theta_i)\cdot c_{\BK}(\x, \y).
    \label{eq:sem-loss}
\end{equation*}
When $\BK$ is compiled into a constraint circuit $c_{\BK}$ one retrieves $-\mathcal{Z}(\x)$ for a two-circuit version of \semprola that is as expressive as FIL as it assumes independent labels via a conditional PC $\prod_i q(Y_i; \theta_i)$.
The neuro-symbolic entropy (\textsc{NeSyEnt})~\citep{ahmed2022neuro} extends \semloss by an entropy term that improves (but still does not guarantee) consistency. It still makes the same independence assumptions over labels (D2).

\noindent
\textbf{Consistency layers.}  Approaches ensuring consistency by embedding the constraints into the predictive layer as in \semprolas include MultiplexNet~\citep{Hoernle2021MultiplexNetTF} and HMCCN~\citep{giunchiglia2020coherent}.
MultiplexNet is able to encode only constraints in disjunctive normal form, which is problematic for generality (D4) and efficiency (D6) as  neuro-symbolic SOP tasks involve an intractably large number of clauses -- e.g. our pathfinding experiments involves trillions of clauses.
HMCCN encodes label dependencies as fuzzy relaxation and is the current state-of-the-art model for HMLC~\citep{giunchiglia2020coherent}.
Even its recent extension~\citep{giunchiglia2021multi} is restricted to a certain family of constraints (D4) that can be represented with fuzzy logic.

\noindent
\textbf{Other approaches.}  Other common approaches to neuro-symbolic SOP require to invoke a solver to either obtain the MAP state or to compute (often only approximately) the gradient of the loss~\citep{deshwal2019learning,poganvcic2019differentiation,niepert2021implicit}.  \semprolas have no such requirement.
Some neuro-symbolic approaches~\citep{sarker2021neuro} constrain the outputs of neural networks within complex logical reasoning pipelines to solve tasks harder than neuro-symbolic SOP.
For instance, DeepProblog~\citep{manhaeve2018deepproblog} uses Prolog's backward chaining algorithm for first order logical rules whose probabilistic weights are predicted by the network.
In modern implementations of Problog, grounding a first order program and then compiling it into constraint circuits~\citep{dries2015problog2} produces a conditional circuit akin to those we use in \semprolas, but in which (\textit{i}) only input distributions are parameterized and (\textit{ii}) increasing the parameter count is not considered a straightforward operation.

\section{Experiments}\label{sec:experiments}

We evaluate \semprolas on standard neuro-symbolic SOP benchmarks such as \textit{simple path prediction}, \textit{preference learning}~\citep{xu2018semantic}, \textit{shortest path finding in Warcraft}~\citep{poganvcic2019differentiation} and \textit{HMLC}~\citep{giunchiglia2020coherent}.
We compare \semprolas against several state-of-the-art  loss- and layer-based approaches (\cref{sec:related}) by applying them to the same base neural network architecture as feature extractor $f$.
As we are interested in measuring how close to the ground truth and how safe the predictions of all models are,  we report the percentage of \textsc{Exact} matches of the predicted labels, also called subset accuracy~\citep{tsoumakas2007multi}, and the percentage of \textsc{Consistent} predictions, also called ``Constraint''~\citep{xu2018semantic}.
Note that, like other consistency layers, \semprolas are guaranteed to always output $100\%$ consistent predictions.
Additionally, we report the \textsc{Hamming} score~\citep{tsoumakas2007multi}, mainly to maintain compatibility with previous experimental settings~\citep{xu2018semantic,ahmed2022neuro}.
This metric does not consider consistency of predictions and naturally favors competitors that assume label independence and thus can minimize the per-label cross-entropy~\citep{dembczynski2012label} (\cref{tab:warcraft}).
\cref{app:exp-details} collects all experimental details such as architectures and hyperparameters used for each experiment.

\begin{table}[!t]
    \centering
    \caption{\semprolas outperform competitors in pathfinding in Warcraft.
    Predicted paths that do not exactly match the ground truth are still valid paths and yield very close costs to the ground truth. Competitors' predictions can have higher Hamming scores but be invalid.  More examples in \cref{app:warcraft-paths}.
    }
    \label{tab:warcraft}
    \hspace{-10pt}\begin{minipage}{.5\columnwidth}
        \scriptsize\sc\vspace{-5pt}
    \begin{tabular}{l c c c}
    \toprule
    Architecture & Exact & Hamming & Consistent\\
    \midrule
    ResNet-18+FIL &   55.0 & \textbf{97.7}  & 56.9\\
    ResNet-18+\semloss& 59.4 &  \textbf{97.7} & 61.2\\
    ResNet-18+\semprola& \textbf{78.2}  & 96.3  & \textbf{100.0}\\
    \bottomrule
    \end{tabular}
    \end{minipage}\begin{minipage}{.45\columnwidth}
     \centering
    \setlength{\tabcolsep}{2pt}
    \begin{tabular}{cccc}
        \includegraphics[width=0.24\linewidth]{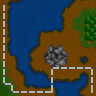}
        & \includegraphics[width=0.24\linewidth]{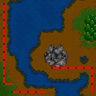}
        & \includegraphics[width=0.24\linewidth]{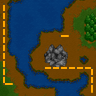}
        & \includegraphics[width=0.24\linewidth]{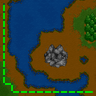}
        \\
        {\scriptsize\textsc{Ground Truth}} & {\scriptsize\textsc{FIL}} & {\scriptsize\semloss} & 
        {\scriptsize\semprola}
    \end{tabular}    
    \end{minipage}
\end{table}

\upparagraph{Simple path prediction \& preference learning.}\label{sec:grids}
We start by comparing \semprolas against loss-based approaches, reproducing the neuro-symbolic benchmarks of~\citet{xu2018semantic} for simple path prediction and preference learning.
In the first experiment, given a source and destination node in an unweighted grid $G = (V, E)$, the neural net needs to find the shortest unweighted path connecting them. 
We consider a $4 \times 4$ grid.
The input $(\x, \y)$ is a binary vector of length $|V|+|E|$, with the first $|V|$ variables indicating the source and destination nodes, and the subsequent $|E|$ variables indicating a subgraph $G' \subseteq G$. Each label is a binary vector of length $|E|$ encoding the unique shortest path in $G'$.
For each example, we obtain $G'$ by dropping one third of the edges in the graph $G$ uniformly at random, filter out the connected components with fewer than $5$ nodes, to reduce degenerate cases, and then sample a source and destination node uniformly at random from $G'$.  The dataset consists of $1600$ such examples, with a $60/20/20$ train/validation/test split.

In the preference learning task, given a user's ranking over a subset of items, the network has to predict the user's ranking over the remaining items. 
We encode an ordering over $n$ items as a binary matrix $Y_{ij}$, where for each $i, j \in 1, \ldots, n$, $Y_{ij}$ indicates whether item $i$ is the $j$th element in the ordering. 
The input $\x$ consist of the user's preference over $6$ sushi types, and the model has to predict the user’s preferences (a strict total order) over the remaining $4$.
We use preference ranking data over $10$ types of sushi for $5,000$ individuals, taken from~\citep{mattei2013preflib}, and a $60$/$20$/$20$ split.

We employ a 5-layer and 3-layer MLP as a baseline for the simple path prediction, and preference learning, respectively, equipped with FIL layer and additionally with the Semantic Loss~\citep{xu2018semantic} (MLP+\semloss) or its entropic extension~\citep{ahmed2022neuro} (MLP+\textsc{NeSyEnt}).
We compile the logical constraints into an SDD~\citep{darwiche2011sdd} and then turn it into a the same constraint circuit $c_{\BK}$  that is used for \semloss, \textsc{NeSyEnt} (\cref{sec:related}) and our 1-circuit implementation of \semprolas.
Table~\ref{tab:sp} clearly shows that the increased expressiveness of \semprola, coming from overparameterizing  $c_{\BK}$, allows to outperform all competitors while guaranteeing consistent predictions, as expected.

\begin{table}[!t]
    \setlength{\tabcolsep}{4pt}
    \centering
    \tiny
    \begin{minipage}{.35\textwidth}
        \caption{Comparison between \semprola and HMCNN~\citep{giunchiglia2020coherent} on twelve HMLC datasets averaged over $10$ runs. Best results for each dataset are in bold. Results which are not significantly worse than the competition, as determined using an unpaired Wilcoxon test, are marked in boldface. Consistency is always $100\%$ for both approaches.}
    \end{minipage}\hfill\begin{minipage}{.6\textwidth}
    \sc
    \begin{tabular}{lrrrrrrrr}
        \toprule
        Dataset
            & \multicolumn{2}{c}{Exact Match}
            & \multicolumn{2}{c}{Hamming Score}
        \\
        \cmidrule(lr){2-3} \cmidrule(lr){4-5}
            & \multicolumn{1}{c}{HMCNN}
            & \multicolumn{1}{c}{MLP+\semprola}
            & \multicolumn{1}{c}{HMCNN}
            & \multicolumn{1}{c}{MLP+\semprola}
        \\
        \midrule
        CellCycle
            & $3.05 \pm 0.11$
            & $\mathbf{3.79 \pm 0.18}$
            & $\mathbf{98.26 \pm 0.00}$
            & $97.84 \pm 0.06$
        \\
        Derisi
            & $1.39 \pm 0.47$
            & $\mathbf{2.28 \pm 0.23}$
            & $\mathbf{98.32 \pm 0.32}$
            & $97.70 \pm 0.07$
        \\
        Eisen
            & $5.40 \pm 0.15$
            & $\mathbf{6.18 \pm 0.33}$
            & $\mathbf{98.09 \pm 0.01}$
            & $97.30 \pm 0.04$
        \\
        Expr
            & $4.20 \pm 0.21$
            & $\mathbf{5.54 \pm 0.36}$
            & $\mathbf{98.29 \pm 0.01}$
            & $97.87 \pm 0.02$
        \\
        Gasch1
            & $3.48 \pm 0.96$
            & $\mathbf{4.65 \pm 0.30}$
            & $\mathbf{98.37 \pm 0.31}$
            & $97.59 \pm 0.05$
        \\
        Gasch2
            & $3.11 \pm 0.08$
            & $\mathbf{3.95 \pm 0.28}$
            & $\mathbf{98.27 \pm 0.00}$
            & $97.94 \pm 0.07$
        \\
        Seq
            & $5.24 \pm 0.27$
            & $\mathbf{7.98 \pm 0.28}$
            & $\mathbf{98.31 \pm 0.01}$
            & $97.66 \pm 0.03$
        \\
        Spo  
            & $\mathbf{1.97 \pm 0.06}$
            & $\mathbf{1.92 \pm 0.11}$
            & $\mathbf{98.23 \pm 0.00}$
            & $98.17 \pm 0.03$
        \\
        Diatoms  
            & $48.21 \pm 0.57$
            & $\mathbf{58.71 \pm 0.68}$
            & $\mathbf{99.75 \pm 0.00}$
            & $99.64 \pm 0.01$
        \\
        Enron  
            & $5.97 \pm 0.56$
            & $\mathbf{8.18 \pm 0.68}$
            & $\mathbf{94.10 \pm 0.04}$
            & $93.19 \pm 0.13$
        \\
        Imclef07a  
            & $79.75 \pm 0.38$
            & $\mathbf{86.08 \pm 0.45}$
            & $\mathbf{99.40 \pm 0.01}$
            & $99.35 \pm 0.03$
        \\
        Imclef07d 
            & $76.47 \pm 0.35$
            & $\mathbf{81.06 \pm 0.68}$
            & $\mathbf{98.06 \pm 0.02}$
            & $\mathbf{98.07 \pm 0.08}$
        \\
        \bottomrule
    \end{tabular}%
    \end{minipage}
    \label{table:hmlc-exp}
\end{table}

\upparagraph{Warcraft Shortest Path.}\label{sec:warcraft}

Next, we evaluate \semprola on the more challenging task of predicting the minimum cost path in a weighted $12 \times 12$ grid imposed over terrain maps of Warcraft II~\citep{poganvcic2019differentiation}.
Each vertex is assigned a cost corresponding to the type of the underlying terrain (e.g., earth has lower cost than water).
The minimum cost path between the top left and the bottom right vertices of the grid is encoded as an indicator matrix, and serves as a label.
As in~\citep{poganvcic2019differentiation} we use a ResNet18~\citep{he2016deep} with FIL optionally with $\semloss$ as a baseline. 
Given the largest size of the compiled constraint circuit $c_{\BK}$ in this case $10^{10}$, we use a two-circuit implementation of \semprola.
Results in \cref{fig:illustration} and \cref{tab:warcraft} are striking: not only \semprola outperforms competitors by a large margin -- approx. $+23$\% over FIL and $+19$\% over the SL -- but also consistently delivers meaningful paths that are very close to the ground truth in terms of cost, even when they encode very different routes. See \cref{app:warcraft-paths} for a gallery of these examples.

\upparagraph{Hierarchical Multi-Label Classification.}\label{sec:hmlc}

Lastly, we follow the experimental setup of~\citet{giunchiglia2020coherent} and evaluate \semprola on $12$ real-world HMLC
tasks spanning four different domains: $8$ functional genomics, $2$ medical images, $1$ microalgea classification, and $1$ text categorization. 
\cref{fig:semprola-circuits} shows an example of a hierarchy of classes.
These tasks are especially challenging due to the limited number of training samples, the large number of output classes, ranging from $56$ to $4130$, as well as the sparsity of the output space.
For numeric features we replaced missing values by their mean, and for categorical features by a vector of zeros, and standardized all features. We used the validation splits to determine the number of layers in the gating function as well as the overparameterization, keeping all other hyperparameters fixed. The final models were obtained by training using a batch size of $128$ and early stopping on the validation set.
We compare our single-circuit \semprola against HMCNN which was shown to outperform several other
state-of-the-art HMLC approaches in \citet{giunchiglia2020coherent}. 
We show the effect of increasing the expressivenss of \semprola via overparameterization in an ablation test in \cref{app:hmlc-overparameterize}.
The results in Table~\ref{table:hmlc-exp} highlight that \semprola significantly outperforms HMCNN in terms of exact match on $11$ data sets and performs comparably on $1$, all while achieving nearly identical Hamming score.

\section{Conclusion}
\label{sec:conclusion}

Our \semprolas offer the first clear interface for integrating complex probabilitistic reasoning and logical constraints on top of any neural network classifier while retaining efficient inference and training.
They improve by a noticeable margin the current state-of-the-art on challenging neuro-symbolic SOP benchmarks such as pathfinding and HMLC.
This opens up a number of interesting research directions.
First, \semprolas can be extended to incorporate logical constraints over multiple networks and representable by first-order formulas~\citep{manhaeve2018deepproblog}, which we plan to explore in future works, making the circuit construction pipeline totally transparent to users~\citep{AhmedAAAI22} while possibly automatically learning constraints from data~\citep{de2018learning,morettin2020learning}.
Second, we are interested in leveraging \semprolas to inject scalable logical constraints into large language models~\citep{bommasani2021opportunities} thus equipping them with robust probabilistic reasoning capabilities~\citep{geva2020injecting,zhang2021greaselm}.
Third, we plan to leverage the benefits of \semprolas for improving efficiency and reliability of human-in-the-loop machine learning tasks~\citep{teso2022efficient}.

\scalebox{0.01}{``La guerra piu totale'' R. Benson}\vspace{-25pt}

\begin{ack}
The authors would like to thank Arthur Choi for helpful discussions on compiling the constraints for the Warcraft Shortest Path task, and Andreas Grivas for proofreading a draft manuscript.
The research of ST was partially supported by TAILOR, a project funded by EU Horizon 2020 research and innovation programme under GA No 952215.
This work is partially supported by a DARPA PTG grant, NSF grants \#IIS-1943641, \#IIS-1956441, \#CCF-1837129, Samsung, CISCO, and a Sloan Fellowship.
\end{ack}

\bibliographystyle{apalike}
\bibliography{referomnia}

\newpage
\appendix

\section{Proofs}
\label{app:proofs}

\textbf{Theorem 3.1} (Efficient inference in \semprolas). \textit{If $q(\Y; \bTheta)$ and $c_{\BK}(\Y, \X)$ are two smooth, decomposable and compatible circuits, then computing \cref{eq:semprola} can be done in $\mathcal{O}(\abs{q}\abs{c})$ time, where $|\cdot|$ denotes the circuit size.
Furthermore, if they are also deterministic, then computing the MAP state can be done in $\mathcal{O}(\abs{q}\abs{c})$ time. }.

We prove the first statement by first showing that the partition function $\mathcal{Z}(\x)$ in \cref{eq:semprola} can solved exactly in time  $\mathcal{O}(\abs{q}\abs{c})$.
It will then follow from it that computing \cref{eq:semprola} can be done in $\mathcal{O}(\abs{q}\abs{c} + \abs{q} + \abs{c})\approx\mathcal{O}(\abs{q}\abs{c})$ where the last two additive factors derive from evaluating $q$ and $c$ for an input configuration $(\x, \y)$.

To do so, we will exploit two ingredients: i) the product of $q$ and $c$  can be represented as a smooth and decomposable circuit in time $\mathcal{O}(\abs{q}\abs{c})$ \citep{vergari2021compositional} and ii) any smooth and decomposable circuit guarantees tractable marginalization in time linear in its size \citep{choi2020pc}.
The next two propositions formalize these statements.

\begin{proposition}[Tractable product of circuits]
\label{prop:trac-product}
Let $q(\Y; \bTheta)$ and $c_{\BK}(\Y, \X)$ be two smooth, decomposable circuits that are compatible over $\Y$ then computing their product  as a circuit $r_{\bTheta, \BK}(\X, \Y)=q(\Y; \bTheta)\cdot c_{\BK}(\Y, \X)$ that is decomposable over $\Y$ can be done in $\mathcal{O}(\abs{q}\abs{c})$.
If both $q$ and $c$ are also deterministic, then $r$ is as well.
\begin{proof}
The proof directly follows from Theorem 3.2 from \citet{vergari2021compositional}.
\end{proof}
\end{proposition}

Note that $\mathcal{O}(\abs{q}\abs{c})$ is a loose upperbound and the size of $r$ is in practice smaller~\citep{vergari2021compositional}. 
%

\begin{proposition}[Tractable marginalization of circuits]
\label{prop:trac-integration}
Let $r(\X,\Y)$ be a circuit that is smooth and decomposable over $\Y$ with input functions over $\Y$ that can be tractably marginalized out. 
Then for any variables $\Y'\subseteq\Y$ and their assignment $\y'$, the marginalization $\sum_{\y'}r(\y',\y'',\x)$ can be computed exactly in time linear in the size of $r$, where $\Y''=\Y\setminus{\Y'}$.
\begin{proof}
The proof follows by considering that i) the input functionals in \semprolas are simple distributions such as Bernoullis and indicators and can be easily marginalized in $\mathcal{O}(1)$ and ii) that for every configuration $\x$ of variables $\X$, $r(\Y, \x)$ is a circuit only over $\Y$ and therefore Proposition 2.1 from \citet{vergari2021compositional} can be directly applied. 
\end{proof}
\end{proposition}

Analogously, the second statement of \cref{thm:prod-pcs} follows from \cref{prop:trac-product} and by recalling that the MAP state of a deterministic circuit can be computed in time linear in its size.

\begin{proposition}[Tractable MAP state of circuits (\citet{choi2020pc})]
\label{prop:trac-map}
Let $r(\X,\Y)$ be a circuit that is smooth and decomposable and deterministic over $\Y$ then for a configuration $\x$ its MAP state $\argmax_{\y} r(\x, \y)$ can be computed in time $\mathcal{O}(\abs{r})$. 
\end{proposition}

\section{Compiling logical formulas into circuits}
\label{app:compilation}
For our experiments we use standard compilation tools to obtain a constraint circuit starting from a propositional logical formula in conjunctive normal form.
Specifically, we use Graphillion\footnote{https://github.com/takemaru/graphillion} to compile the constraints in the Warcraft pathfinding experiment into an SDD.
For all other experiments, we use PySDD\footnote{https://github.com/wannesm/PySDD} a python SDD compiler~\citep{darwiche2011sdd,choi2013dynamic}.

We now illustrate step-by-step one example of such a compilation for a simple logical formula.
Consider the constraint circuit $c$ in \cref{fig:semprola-circuits} encoding the constraint
\begin{equation}
    (Y_{\mathsf{cat}} \implies Y_{\mathsf{animal}}) \wedge (Y_{\mathsf{dog}} \implies Y_{\mathsf{animal}}).
\end{equation}
Intuitively, our aim is to compile the above logical formula into a \emph{compact}
form representing all possible assignments to  $Y_{\mathsf{cat}}, Y_{\mathsf{dog}},
Y_{\mathsf{animal}}$ satisfying the above constraint.
We compile such a constraint by proceeding in a bottom up fashion,
where bottom-up compilation can be seen as composing Boolean sub-functions whose domain is determined by a variable ordering, also called vtree (see \cref{sec:efficient-inference-semprolas}).
In this example, we assume the function $f(Y_{\mathsf{animal}}, Y_{\mathsf{cat}}, Y_{\mathsf{dog}})$ decomposes as $f_1(Y_{\mathsf{animal}}) \cdot f_2(Y_{\mathsf{dog}}) \cdot f_3(Y_{\mathsf{cat}})$.
We therefore start by compiling a constraint circuit that is a function of $Y_{\mathsf{cat}}$ and $Y_{\mathsf{dog}}$, and compose it with a constraint circuit that is a function of $Y_{\mathsf{animal}}$.
We first introduce input functionals representing indicators associated with $Y_{\mathsf{cat}},Y_{\mathsf{dog}}, Y_{\mathsf{animal}}$.
We will denote by $Y_{i}$ the indicator $\Ind{Y_{i}=1}$ and by $\lnot Y_{i}$ the indicator $\Ind{Y_{i}=0}$.\\[10pt]
    \includegraphics[width=.15\columnwidth,page=9]{figs/semprola.pdf}\hfill\includegraphics[width=.15\columnwidth,page=10]{figs/semprola.pdf}\hfill \includegraphics[width=.15\columnwidth,page=11]{figs/semprola.pdf}\hfill \includegraphics[width=.15\columnwidth,page=12]{figs/semprola.pdf}\hfill \includegraphics[width=.15\columnwidth,page=13]{figs/semprola.pdf}\hfill \includegraphics[width=.15\columnwidth,page=14]{figs/semprola.pdf}

%
We start by disjoining the indicators $Y_{\mathsf{cat}}$ with $\lnot Y_{\mathsf{cat}}$, and $Y_{\mathsf{dog}}$ with $\lnot Y_{\mathsf{dog}}$.
This corresponds to introducing deterministic and smooth sum units in our circuits.\\[10pt]
    \hspace{100pt}\includegraphics[width=.2\columnwidth,page=15]{figs/semprola.pdf}\hspace{100pt}\includegraphics[width=.2\columnwidth,page=16]{figs/semprola.pdf}\hfill

These units represent \emph{disjoint solutions} to the logical formula, meaning there exists distinct assignments, characterized by the children, that satisfy the logical constraint e.g. $Y_{\mathsf{cat}}, Y_{\mathsf{dog}}, Y_{\mathsf{animal}}$ and $Y_{\mathsf{cat}}, \lnot Y_{\mathsf{dog}}, Y_{\mathsf{animal}}$ are two distinct assignments that satisfy the logical constraint.

The compilation process proceeds by conjoining the constraint circuits for $Y_{\mathsf{dog}} \lor \lnot Y_{\mathsf{dog}}$ with $Y_{\mathsf{cat}}$, $Y_{\mathsf{dog}}$ with $Y_{\mathsf{cat}} \lor \lnot Y_{\mathsf{cat}}$, and $\lnot Y_{\mathsf{dog}}$ with $\lnot Y_{\mathsf{cat}}$.\\[10pt]
    \hspace{100pt}\includegraphics[width=.3\columnwidth,page=17]{figs/semprola.pdf}
    
A decomposable product units \emph{composes} functions over {disjoint sets of variables}. The above three product nodes represent the Boolean functions $(Y_{\mathsf{dog}} \lor \lnot Y_{\mathsf{dog}}) \land Y_{\mathsf{cat}}$, $Y_{\mathsf{dog}} \land (Y_{\mathsf{cat}} \lor \lnot Y_{\mathsf{cat}})$, and  $\lnot Y_{\mathsf{dog}} \land \lnot Y_{\mathsf{cat}}$.

We again disjoin $(Y_{\mathsf{dog}} \lor \lnot Y_{\mathsf{dog}}) \land Y_{\mathsf{cat}}$ with $Y_{\mathsf{dog}} \land (Y_{\mathsf{cat}} \lor \lnot Y_{\mathsf{cat}})$, and $\lnot Y_{\mathsf{dog}} \land \lnot Y_{\mathsf{cat}}$ with $\mathsf{true}$, the logical multiplicative identity, guaranteeing alternating sum and product nodes, as mentioned in \cref{sec:prob-circuits}.

So far, we have compiled constraint circuits for the logical formula
\begin{equation}
\label{eq:c1}
    ((Y_{\mathsf{dog}} \lor \lnot Y_{\mathsf{dog}}) \land Y_{\mathsf{cat}}) \lor (Y_{\mathsf{dog}} \land (Y_{\mathsf{cat}} \lor \lnot Y_{\mathsf{cat}}))
\end{equation}
and the logical formula
\begin{equation}
\label{eq:c2}
    \lnot Y_{\mathsf{dog}} \land \lnot Y_{\mathsf{cat}}
\end{equation}

\includegraphics[width=.4\columnwidth,page=18]{figs/semprola.pdf}

What remains is to conjoin \cref{eq:c1} with $Y_{\mathsf{animal}}$, and \cref{eq:c2} with $\lnot Y_{\mathsf{animal}}$, and disjoin the resulting constraint circuits.
What we get is a a mixture distribution over the possible solutions of the constraint:
If we predict there is a dog or a cat, or both, in e.g., an image, we better predict that there's an animal.
On the other hand, the absence of a dog and a cat from an image implies nothing as to the presence of an animal in the image.

\includegraphics[width=.65\columnwidth,page=19]{figs/semprola.pdf}

Compilation techniques like the one we illustrated do not, however, escape the hardness of 
the problem: the compiled circuit can be
exponential in the size of the constraint, \textit{in the worst case}.
\textit{In practice}, nevertheless, we can obtain compact circuits because real-life  logical constraints exhibit enough structure (e.g., they encode repeated
sub-problems) that can be easily exploited by a compiler.
We refer to the
literature of compilation for details on this~\citep{darwiche2002knowledge}.

\section{Overparameterize the single-circuit \semprola}
\label{app:overparameterize}
As mentioned in \cref{def:single-circuit-semprola}, \semprolas can be realized as a single circuit by first compiling a complex logical constraint into a deterministic constraint
circuit, and then parameterizing it using a gating function of the network embeddings.
Intuitively, this parameterization induces a probability distribution over the possible solutions of a logical formula encoded in the constraint circuit.
The expressiveness of this distribution depends on the number of parameters of the constraint circuit, i.e., the number of weighted edges associated to sum units.
As we would like to endow our single-circuit \semprola with the ability to induce complex distributions, we devise \textit{two strategies} to introduce more parameters than what the constraint circuit alone can offer: \textit{replication} and \textit{mixture multiplication}.
Replication works by maintaining $m$ copies of the circuit, and taking their weighted average, i.e., introducing a sum unit that mixes them~\citep{peharz2020random}.
Mixture multiplication, instead, substitutes a single local marginal distribution encoded by a sub-circuit rooted into a sum unit with $k$ mixture models over the same scope. 
In practice, we create $k-1$ copies of each sum units and rewire them by computing a cross product of their inputs as in \citet{peharz2020random}.
\cref{alg:overparam} formalizes this process.

As mentioned in \cref{def:single-circuit-semprola}, both strategies relax determinism.
However, note that \textit{they do not alter the support of the underlying distribution}.
This guarantees that all the predictions will be consistent with the encoded constraint (D3)~(\cref{sec:design}).

\begin{algorithm}[tb]
   \caption{$\overparam(c, k, \mathsf{cache}, \mathsf{first\_call}$)}
   \label{alg:overparam}
   \begin{algorithmic}[1]
   \STATE {\bfseries Input:} a smooth, deterministic, and structured-decomposable circuit $c$ over variables $\X$, an overparameterization factor $k$, and a cache for memoization, and a flag to denote the first call
   \STATE {\bfseries Output:} an overparameterized, smooth, and structured-decomposable circuit $c$ over $\X$
  \IF{$\q \in \mathsf{cache}$}
    \STATE \textbf{return} $\mathsf{cache}\left[\q\right]$
  \ENDIF
  \IF{$c$ is an input unit}
      \STATE $\mathsf{nodes}\leftarrow\left[c\right]$
      \ELSIF{$c$ is a sum unit}
      \STATE $\mathsf{elements} \leftarrow [\;]$
      \STATE \textcolor{ourspecialtextcolor}{//For every product unit that is an input of $c$}
      \STATE \textcolor{ourspecialtextcolor}{//recursively overparameterize its inputs,}
      \STATE \textcolor{ourspecialtextcolor}{//which are sum units, and take their cross (cartesian) product}
      \FOR{$(c_L, c_R) \in \ch(c)$}
          \STATE $\mathsf{left} \leftarrow  \overparam(c_L, k)$
          \STATE $\mathsf{right} \leftarrow  \overparam(c_R, k)$
          \STATE $\mathsf{elements}.\textsc{append}([\textsc{CrossProduct}(\mathsf{left}, \mathsf{right})]$
      \ENDFOR
      \STATE $\ch(c) \leftarrow \mathsf{elements}$
      \STATE $\mathsf{nodes} = [c] + [\textsc{copy}(c) \textbf{ for } i = 1 \textbf{ to } k]$
  \ENDIF
  \IF{$\mathsf{first\_call}$}
  \STATE \textcolor{ourspecialtextcolor}{//Create a sum unit whose inputs are $\mathsf{nodes}$}
  \STATE \textcolor{ourspecialtextcolor}{//and whose parameters are $1$s.}
  \STATE $\mathsf{nodes} \leftarrow \sumUnit(\mathsf{nodes}, \{1\}_{i=1}^{|\mathsf{nodes}|})$

  \ENDIF
  \STATE $\mathsf{cache}(c)\leftarrow \mathsf{nodes}$
   \STATE \textbf{return} $\mathsf{nodes}$
\end{algorithmic}
\end{algorithm}
\section{Additional experimental details}

\label{app:exp-details}

\subsection{Simple path prediction and preference learning}
In the simple path prediction task, given a source and destination node in an unweighted grid $G = (V, E)$, the neural net needs to find the shortest unweighted path connecting them. 
We consider a $4 \times 4$ grid.
The input $(\x, \y)$ is a binary vector of length $|V|+|E|$, with the first $|V|$ variables indicating the source and destination nodes, and the subsequent $|E|$ variables indicating a subgraph $G' \subseteq G$. Each label is a binary vector of length $|E|$ encoding the unique shortest path in $G'$.
For each example, we obtain $G'$ 
by dropping one third of the edges in the graph $G$ uniformly at random, filtering out the connected components with fewer than $5$ nodes, to reduce degenerate cases, and then sample a source and destination node uniformly at random from $G'$. 
The dataset consists of $1600$ such examples, with a $60/20/20$ train/validation/test split.

In the preference learning task, given a user's ranking over a subset of items, the network has to predict the user's ranking over the remaining items. 
We encode an ordering over $n$ items as a binary matrix $Y_{ij}$, where for each $i, j \in 1, \ldots, n$, $Y_{ij}$ indicates whether item $i$ is the $j$th element in the ordering.
The input $\x$ consist of the user's preference over $6$ sushi types, and the model has to predict the user’s preferences (a strict total order) over the remaining $4$.
We use preference ranking data over $10$ types of sushi for $5,000$ individuals, taken from~\citep{mattei2013preflib}, and a $60$/$20$/$20$ split.

We follow~\citet{xu2018semantic} in employing a $5$-layer with $50$ hidden units each and sigmoid activation functions, and 3-layer MLP with $50$ hidden units each as a baseline for the simple path prediction, and preference learning, respectively.
We equip this baselines with a FIL and additionally with the Semantic Loss~\citep{xu2018semantic} (MLP+\semloss) or its entropic extension~\citep{ahmed2022neuro} (MLP+\textsc{NeSyEnt}).

We compile the logical constraints into an SDD~\citep{darwiche2011sdd} and then turn it into a constraint circuit $c_{\BK}$  that is used for \semloss, \textsc{NeSyEnt} (\cref{sec:related}) and our 1-circuit implementation of \semprolas.
To obtain the results for \semprola in Table~\ref{tab:sp}, we perform a grid search over the using the validation set for a maximum of $2000$ iterations, similar to~\citet{xu2018semantic}.
We search over the learning rates in the range $\{1\times10^{-3}, 5\times10^{-3}, 1\times10^{-4}, 5\times10^{-4}\}$, the overparameterization factor $k$ in the range $\{2, 4, 8\}$, as well as the number of circuit mixtures $m$ in the range $\{2,4,8\}$, evaluating the model with the best performance on the validation set.

\subsection{Hierarchical Multi-Label Classification}
We follow the experimental setup of~\citet{giunchiglia2020coherent} and evaluate \semprola on $12$ real-world HMLC
tasks spanning four different domains: $8$ functional genomics, $2$ medical images, $1$ microalgea classification, and $1$ text categorization. 
These tasks are especially challenging due to the limited number of training samples, the large number of output classes, ranging from $56$ to $4130$, as well as the sparsity of the output space.
We used the same train-validation-test splits and experimental setup as \citep{giunchiglia2020coherent}.  
For numeric features we replaced missing values by their mean, and for categorical features by a vector of zeros, and standardized all features. We used the validation splits to determine the number of layers in the gating function in the range $\{2, 4, 8\}$, the overparameterization factor in the range $\{2, 4, 8\}$, and the number of mixtures in the range $\{2, 4, 8\}$, keeping all other hyperparameters fixed. The final models were obtained by training using a batch size of $128$ and early stopping with a patience of $20$ on the validation set.

\subsection{Warcraft pathfinding}\label{app:warcraft-paths}
We evaluate \semprola on the more challenging task of predicting the minimum cost path in a weighted $12 \times 12$ grid imposed over terrain maps of Warcraft II~\citep{poganvcic2019differentiation}.
Our setting differs from the one proposed by \citet{poganvcic2019differentiation} in two ways: i) a node only neighbors four nodes as instead of eight, excluding the diagonals; ii) the neural network predicts the edges in the path, as opposed to the vertices, resolving ambiguities in the previous task (note that a set of vertices can \emph{might} ambiguously encode more than one path).
Each vertex is assigned a cost corresponding to the type of the underlying terrain (e.g., earth has lower cost than water).
The minimum cost path between the top left and the bottom right vertices of the grid is encoded as an indicator matrix, and serves as a label.

We use Graphillion\footnote{https://github.com/takemaru/graphillion} to compile the path constraint, limiting our constraint to the set of paths whose length is less than $29$, as determined on the training set.

As in~\citep{poganvcic2019differentiation} we use a ResNet18~\citep{he2016deep} with FIL optionally with $\semloss$ as a baseline. 
Given the largest size of the compiled constraint circuit $c_{\BK}$ in this case $10^{10}$, we use a two-circuit implementation of \semprola.
We use the identity function as our gating function and do a grid search over only the number of mixtures in the range $\{2, 4, 8\}$ in our model,
keeping all other hyperparameters as proposed in~\citep{poganvcic2019differentiation}.

\begin{figure}[!t]
    \centering
    \setlength{\tabcolsep}{5pt}
    \begin{tabular}{cccc}
            \textsc{Ground Truth} & \textsc{ResNet-18} & \textsc{Semantic Loss} & 
        \semprola (ours)\\
        \includegraphics[width=0.22\linewidth]{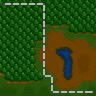}
        & \includegraphics[width=0.22\linewidth]{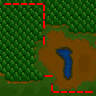}
        & \includegraphics[width=0.22\linewidth]{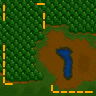}
        & \includegraphics[width=0.22\linewidth]{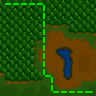}\\
        cost: $55.22$ & cost:$\infty$ & cost:$\infty$ & cost: $55.22$\\[5pt]
        \includegraphics[width=0.22\linewidth]{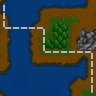}
        & \includegraphics[width=0.22\linewidth]{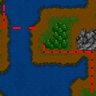}
        & \includegraphics[width=0.22\linewidth]{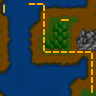}
        & \includegraphics[width=0.22\linewidth]{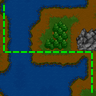}\\
        cost: $57.31$ & cost:$\infty$ & cost:$\infty$ & cost: $58.09$
        \\[5pt]
        \includegraphics[width=0.22\linewidth]{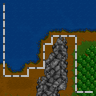}
        & \includegraphics[width=0.22\linewidth]{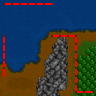}
        & \includegraphics[width=0.22\linewidth]{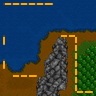}
        & \includegraphics[width=0.22\linewidth]{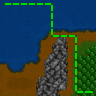}\\
        cost: $97.38$ & cost:$\infty$ & cost:$\infty$ & cost: $98.38$
        \\[5pt]
        \includegraphics[width=0.22\linewidth]{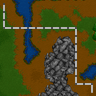}
        & \includegraphics[width=0.22\linewidth]{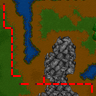}
        & \includegraphics[width=0.22\linewidth]{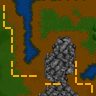}
        & \includegraphics[width=0.22\linewidth]{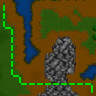}\\
        cost: $30.50$ & cost:$\infty$ & cost:$\infty$ & cost: $30.80$
        \\[5pt]
        \includegraphics[width=0.22\linewidth]{figs/warcraft_images/570_gt.png}
        & \includegraphics[width=0.22\linewidth]{figs/warcraft_images/570_base.png}
        & \includegraphics[width=0.22\linewidth]{figs/warcraft_images/570_sl.png}
        & \includegraphics[width=0.22\linewidth]{figs/warcraft_images/570_op.png}\\
        cost: $39.31$ & cost:$\infty$ & cost:$\infty$ & cost: $45.09$
        \\[5pt]
    \end{tabular}
    \caption{
    \textbf{More examples of shortest path predictions in \semprolas and competitors.}
    \semprolas always deliver valid paths and even when these do not exact match the ground truth, they are very close in terms of their global cost.
    Paths from the baselines might yield a higher Hamming score (as they have more overlapping edges with the ground truth) but are invalid.} 
    \label{fig:more-warcraft}
\end{figure}

\subsection{A study on the effect of overparameterization in \semprola}
We now illustrate the effect that overparameterization has on the performance of the single-circuit \semprola.
To that end, we performed an ablation study, comparing single-circuit \semprolas comprising a different number of
circuit copies $m$ for our replication strategy, a different number of layers in the gating function, denoted by \emph{Gates}, and
the overparameterization factor $k$ as used in \cref{alg:overparam} in our mixture multiplication strategy.

We report the exact match percentage of the predicted labels on the validation set of the 12 HMLC datasets in 
\cref{table:hmlc-exp-app}.
As a general trend, we can see that our overparameterization strategies pay off and in general more mixture nodes help ($k=4$) as well as using more replicas ($m\geq 4$).
The effect of employing a deeper gating function is less striking instead, with a two-layer gating function achieving highest performances on 9 datasets.
Our recommendation 

\begin{table}[!t]
    \caption{A comparison of the performance of single-circuit \semprola with different parameters: $m$, the number of circuit copies in our replication strategy; \emph{gates}, the number of layers in the gating function; and $k$ the overparameterization factor in the mixture multiplication strategy (\cref{alg:overparam}). We report the percentage of exact matches of the predicted labels on the validation set of the \textit{HMLC} datasets, highlighting the best numbers in \textbf{boldface}. As can be seen, all datasets benefit from overparameterization.}
    \label{app:hmlc-overparameterize}
    \setlength{\tabcolsep}{3.5pt}
    \begin{minipage}{1.0\textwidth}
    \centering
    \sc
    \small
    \begin{tabular}{lrrrrrrrrrrrr}
        \toprule
        Dataset
            & \multicolumn{4}{c}{$m$: $2$}
            & \multicolumn{4}{c}{$m$: $4$}
            & \multicolumn{4}{c}{$m$: $8$}
        \\
            \cmidrule(lr){2-5} \cmidrule(lr){6-9} \cmidrule(lr){10-13}
            & \multicolumn{2}{c}{Gates: $2$}
            & \multicolumn{2}{c}{Gates: $4$}
            & \multicolumn{2}{c}{Gates: $2$}
            & \multicolumn{2}{c}{Gates: $4$}
            & \multicolumn{2}{c}{Gates: $2$}
            & \multicolumn{2}{c}{Gates: $4$}
        \\
        \cmidrule(lr){2-5} \cmidrule(lr){6-9} \cmidrule(lr){10-13}
            & \multicolumn{1}{c}{$k$: $2$}
            & \multicolumn{1}{c}{$k$: $4$}
            & \multicolumn{1}{c}{$k$: $2$}
            & \multicolumn{1}{c}{$k$: $4$}
            & \multicolumn{1}{c}{$k$: $2$}
            & \multicolumn{1}{c}{$k$: $4$}
            & \multicolumn{1}{c}{$k$: $2$}
            & \multicolumn{1}{c}{$k$: $4$}
            & \multicolumn{1}{c}{$k$: $2$}
            & \multicolumn{1}{c}{$k$: $4$}
            & \multicolumn{1}{c}{$k$: $2$}
            & \multicolumn{1}{c}{$k$: $4$}
        \\
        \midrule
        cellcycle & $4.25$ & $4.48$ & $4.48$ & $4.01$ & $4.60$ & $\mathbf{4.83}$ & $4.25$ & $4.48$ & $4.36$ & $4.13$ & $4.36$ & $4.13$\\
        derisi & $2.26$ & $2.02$ & $2.14$ & $2.26$ & $\mathbf{2.49}$ & $2.26$ & $2.38$ & $2.38$ & $\mathbf{2.49}$ & $2.38$ & $2.26$ & $\mathbf{2.49}$\\
        eisen & $6.05$ & $6.05$ & $6.05$ & $6.05$ & $5.86$ & $6.43$ & $\mathbf{6.81}$ & $6.24$ & $6.43$ & $6.43$ & $6.05$ & $6.43$\\
        expr & $5.42$ & $4.83$ & $5.18$ & $5.30$ & $4.83$ & $\mathbf{5.54}$ & $\mathbf{5.54}$ & $5.18$ & $\mathbf{5.54}$ & $5.42$ & $5.18$ & $5.42$\\
        gasch1 & $5.56$ & $5.79$ & $5.67$ & $5.91$ & $5.44$ & $5.67$ & $6.03$ & $\mathbf{6.26}$ & $5.79$ & $5.79$ & $\mathbf{6.26}$ & $6.03$\\
        gasch2 & $4.00$ & $4.24$ & $4.83$ & $\mathbf{4.95}$ & $4.12$ & $4.00$ & $4.12$ & $4.36$ & $4.24$ & $3.53$ & $4.24$ & $4.59$\\
        seq & $7.74$ & $7.74$ & $7.51$ & $7.85$ & $8.19$ & $7.28$ & $7.96$ & $7.17$ & $7.96$ & $7.39$ & $7.51$ & $\mathbf{8.42}$\\
        spo & $2.27$ & $2.15$ & $2.15$ & $2.51$ & $2.39$ & $2.27$ & $2.51$ & $2.51$ & $\mathbf{2.87}$ & $2.27$ & $2.39$ & $2.63$\\
        diatoms & $53.71$ & $\mathbf{54.68}$ & $50.16$ & $51.29$ & $53.23$ & $52.10$ & $49.35$ & $48.23$ & $52.90$ & $52.58$ & $46.61$ & $47.26$\\
        enron & $19.53$ & $18.52$ & $17.85$ & $19.87$ & $19.87$ & $20.20$ & $\mathbf{20.54}$ & $20.20$ & $19.53$ & $20.20$ & $19.53$ & $19.87$\\
        imclef07a & $86.97$ & $87.03$ & $86.27$ & $86.60$ & $87.00$ & $\mathbf{87.33}$ & $86.50$ & $86.70$ & $87.07$ & $86.90$ & $87.00$ & $86.83$\\
        imclef07d & $85.93$ & $85.80$ & $85.87$ & $85.73$ & $85.60$ & $\mathbf{86.50}$ & $85.87$ & $85.90$ & $85.87$ & $85.83$ & $86.10$ & $85.50$
        \\
        \bottomrule
    \end{tabular}%
    \end{minipage}
    \label{table:hmlc-exp-app}
\end{table}

\end{document}